\documentclass[a4paper]{article}

\usepackage{INTERSPEECH2018}

\usepackage[mode=buildnew]{standalone}

\usepackage{microtype}

\usepackage[nopostdot,nonumberlist]{glossaries}
\setacronymstyle{long-short}  
\newacronym{ae}{AE}{autoencoder}
\newacronym{cae}{cAE}{correspondence autoencoder}
\newacronym{mfccs}{MFCCs}{Mel-frequency cepstral coefficients}
\newacronym{plps}{PLPs}{perceptual linear prediction coefficients}
\newacronym{dtw}{DTW}{dynamic time warping}
\newacronym{asr}{ASR}{automatic speech recognition}
\newacronym{utd}{UTD}{unsupervised term discovery}
\newacronym{ap}{AP}{average precision}
\newacronym{sw}{SW}{same-word}
\newacronym{dw}{DW}{different-word}
\newacronym{swsp}{SWSP}{same-word same-speaker}
\newacronym{swdp}{SWDP}{same-word different-speaker}
\newacronym{dwsp}{DWSP}{different-word, same-speaker}
\newacronym{dwdp}{DWDP}{different-word, different-speaker}
\newacronym{gmm}{GMM}{Gaussian mixture model}
\newacronym{hmm}{HMM}{hidden Markov model}
\newacronym{ubm}{UBM}{universal background model}
\newacronym{dnn}{DNN}{deep neural network}
\newacronym{rms}{RMS}{root mean square}
\newacronym{vtln}{VTLN}{vocal tract length normalisation}
\newacronym{lvtln}{LVTLN}{linear VTLN}
\newacronym{zrsc}{ZRSC}{Zero Resource Speech Challenge}
\newacronym{bnfs}{BNFs}{bottleneck features}
\newacronym{cmn}{CMN}{cepstral mean normalization}
\newacronym{cvn}{CVN}{cepstral variance normalization}
\newacronym{cmvn}{CMVN}{cepstral mean and variance normalization}
\newacronym{sat}{SAT}{speaker adaptive training}
\newacronym{wsj}{WSJ}{Wall Street Journal}
\newacronym{dpgmm}{DPGMM}{Dirichlet Process Gaussian mixture model}
\newacronym{tdnn}{TDNN}{time-delay neural network}
\newacronym{sgmm}{SGMM}{subspace Gaussian mixture model}

\usepackage[noadjust]{cite}

\title{Multilingual bottleneck features for subword modeling\\ in zero-resource languages}
\name{Enno Hermann$^1$, Sharon Goldwater$^1$}
\address{
  $^1$ILCC, School of Informatics, University of Edinburgh, UK}
\email{ehermann@inf.ed.ac.uk, sgwater@inf.ed.ac.uk}

\begin{document}

\maketitle

\begin{abstract}
  How can we effectively develop speech technology for languages where no
  transcribed data is available? Many existing approaches use no annotated
  resources at all, yet it makes sense to leverage information from large
  annotated corpora in other languages, for example in the form of multilingual
  \gls{bnfs} obtained from a supervised speech recognition system. In this work,
  we evaluate the benefits of \gls{bnfs} for subword modeling (feature
  extraction) in six unseen languages on a word discrimination task. First we
  establish a strong unsupervised baseline by combining two existing methods:
  \gls{vtln} and the \gls{cae}. We then show that \gls{bnfs} trained on a single
  language already beat this baseline; including up to 10 languages results in
  additional improvements which cannot be matched by just adding more data from
  a single language. Finally, we show that the \gls{cae} can improve further on
  the \gls{bnfs} if high-quality same-word pairs are available.
\end{abstract}

\glsresetall 

\section{Introduction}
Recent years have seen increasing interest in ``zero-resource'' speech
technology: systems developed for a target language without transcribed data or
other hand-curated resources. One challenge for these systems, highlighted by
the \gls{zrsc} of 2015 \cite{Versteegh2015} and 2017 \cite{Dunbar2017}, is to
improve subword modeling, i.e., to extract speech features from the target
language audio that work well for word discrimination or downstream tasks such
as query-by-example.

The ZRSCs were motivated largely by questions in artificial intelligence and
human perceptual learning, and focused on approaches where no transcribed data
from {\em any} language is used. Yet from an engineering perspective it also
makes sense to explore how training data from higher-resource languages can be
used to improve speech features in a zero-resource language.

There is considerable evidence that \gls{bnfs} extracted using a multilingually
trained \gls{dnn} can improve ASR for target languages with just a few hours of
transcribed data \cite{Vesely2012,Vu2012,Thomas2012,Cui2015,Alumae2016}.
However, there has been little work so far exploring supervised multilingual
\gls{bnfs} for target languages with no transcribed data at all.
\cite{Yuan2016,Renshaw2015} trained {\em monolingual} BNF extractors and showed
that applying them cross-lingually improves word discrimination in a
zero-resource setting. \cite{Yuan2017,Chen2017} trained a multilingual \gls{dnn}
to extract BNFs for a zero-resource task, but the \gls{dnn} itself was trained
on untranscribed speech: an unsupervised clustering method was applied to each
language to obtain phone-like units, and the \gls{dnn} was trained on these
unsupervised phone labels.

We know of only two previous studies of supervised multilingual BNFs for
zero-resource speech tasks. In \cite{Yuan2017a}, the authors trained \gls{bnfs}
on either Mandarin, Spanish or both, and used the trained \gls{dnn}s to extract
features from English (simulating a zero-resource language). On a
query-by-example task, they showed that \gls{bnfs} always performed better than
MFCCs, and that bilingual \gls{bnfs} performed as well or better than
monolingual ones. Further improvements were achieved by applying weak
supervision in the target language using a correspondence autoencoder
\cite{Kamper2015} trained on English word pairs. However, the authors did not
experiment with more than two training languages, and only evaluated on English.

In the second study \cite{Shibata2017}, the authors built multilingual systems
using either seven or ten high-resource languages, and evaluated on the three
``development'' and two ``surprise'' languages of the \gls{zrsc} 2017. However,
they included transcribed training data from four out of the five evaluation
languages, so only one language's results (Wolof) are truly zero-resource.

This paper presents a more thorough evaluation of multilingual \gls{bnfs},
trained on between one and ten languages from the GlobalPhone collection and
evaluated on six others. We show that training on more languages consistently
improves performance on word discrimination, and that the improvement is not
simply due to more training data: an equivalent amount of data from one language
fails to give the same benefit.

Since BNF training uses no target language data at all, we also compare to
methods that train unsupervised on the target language, either alone or in
combination with the multilingual training. We use a \gls{cae}
\cite{Kamper2015}, which learns to abstract away from signal noise and
variability by training on pairs of speech segments extracted using an \gls{utd}
system---i.e., pairs that are likely to be instances of the same word or phrase.
In the setting with target language data only, we find that applying \gls{vtln}
to the input of both the \gls{utd} and \gls{cae} systems improves the learned
features considerably, suggesting that \gls{cae} and \gls{vtln} abstract over
different aspects of the signal. Nevertheless, \gls{bnfs} trained on just a
single other language already outperform the \gls{cae}-only training, with
multilingual \gls{bnfs} doing better by a wide margin.

We then tried fine-tuning the multilingual \gls{bnfs} to the target language by
using them as input to the \gls{cae}. When trained with \gls{utd} word pairs, we
found no benefit to this fine-tuning. However, training with manually labeled
word pairs did yield benefits, suggesting that this type of supervision can help
fine-tune the \gls{bnfs} if the word pairs are sufficiently high-quality.

\section{Experimental setup}
\subsection{Dataset}
We use 16~languages from the GlobalPhone corpus of speech read from news
articles~\cite{Schultz2013}. The selected languages and dataset sizes are shown
in Table~\ref{tab:data}. We consider the 10~languages in the top section with a
combined 198.3~hours of speech as high-resource languages, where transcriptions
are available to train a supervised \gls{asr} system. We treat the 6~languages
in the bottom section as zero-resource languages on which we evaluate the new
feature representations.

In addition we use the English \gls{wsj} corpus~\cite{Paul1992} which is
comparable to the GlobalPhone corpus. We either use the entire 81~hours or only
a 15~hour subset, so that we can compare the effect of increasing the
amount of data for one language with training on data from 4~GlobalPhone
languages.

\begin{table}[th]
  \caption{Dataset sizes (hours). About 100~speakers per language
    with 80\% of these in the training set and no speaker overlap.}
  \label{tab:data}
  \centering
  \begin{tabular}{l l c c c}
    \toprule
    \multicolumn{2}{c}{\textbf{Language}} & 
    \multicolumn{1}{c}{\textbf{Train}} &
    \multicolumn{1}{c}{\textbf{Dev}} &
    \multicolumn{1}{c}{\textbf{Test}} \\
    \midrule
    \textit{High-resource} \\
    Bulgarian  & (BG) & 17.1 & 2.3 & 2.0 \\
    Czech      & (CS) & 26.8 & 2.4 & 2.7 \\
    French     & (FR) & 22.8 & 2.1 & 2.0 \\
    German     & (DE) & 14.9 & 2.0 & 1.5 \\
    Korean     & (KO) & 16.6 & 2.2 & 2.1 \\
    Polish     & (PL) & 19.4 & 2.8 & 2.3 \\
    Portuguese & (PT) & 22.8 & 1.6 & 1.8 \\
    Russian    & (RU) & 19.8 & 2.5 & 2.4 \\
    Thai       & (TH) & 21.2 & 1.5 & 0.4 \\
    Vietnamese & (VI) & 16.9 & 1.4 & 1.5 \\
    \midrule
    English81 WSJ & (EN) & 81.3 & 1.1 & 0.7 \\
    English15 WSJ &      & 15.1 &  -  &  -  \\
    \midrule
    \textit{Zero-resource} \\
    Croatian   & (HR) & 12.1 & 2.0 & 1.8 \\
    Hausa      & (HA) &  6.6 & 1.0 & 1.1 \\
    Mandarin   & (ZH) & 26.6 & 2.0 & 2.4 \\
    Spanish    & (ES) & 17.6 & 2.1 & 1.7 \\
    Swedish    & (SV) & 17.4 & 2.1 & 2.2 \\
    Turkish    & (TR) & 13.3 & 2.0 & 1.9 \\
    \bottomrule
  \end{tabular}
\end{table}

\subsection{Baseline features}
For baseline features, we use Kaldi~\cite{Povey2011} to extract
MFCCs+$\Delta$+$\Delta\Delta$ and PLPs+$\Delta$+$\Delta\Delta$ with a window
size of 25~ms and a shift of 10~ms, and we apply per-speaker \acrlong{cmn}. We
also evaluated MFCCs and PLPs with \acrfull{vtln}, a simple feature-space
speaker adaptation technique that normalizes a speaker's speech by warping the
frequency-axis of the spectra. \gls{vtln} models are trained using maximum
likelihood estimation under a given acoustic model---here, a diagonal-covariance
\acrlong{ubm} with 1024~components trained on each language's training data.
Warp factors can then be extracted for both the training and for unseen data.

\subsection{Bottleneck features}
For monolingual training of the high-resource languages, we follow the Kaldi
recipes for the GlobalPhone and WSJ corpora and train a \gls{sgmm} system for
each language to get initial context-dependent state alignments; these states
serve as targets for \gls{dnn} training.

For multilingual training, we closely follow the existing Kaldi recipe for the
Babel corpus. We train a \gls{tdnn}~\cite{Peddinti2015} with block
softmax~\cite{Grezl2014}, i.e. all hidden layers are shared between languages,
but there is a separate output layer for each language and for each training
instance only the error at the corresponding language's output layer is used to
update the weights. The \gls{tdnn} has six 625-dimensional hidden
layers\footnote{The splicing indexes are \texttt{-1,0,1 -1,0,1 -1,0,1 -3,0,3
    -3,0,3 -6,-3,0 0}.} followed by a 39-dimensional bottleneck layer with ReLU
activations and batch normalization. Each language then has its own
625-dimensional affine and a softmax layer. The inputs to the network are
40-dimensional MFCCs with all cepstral coefficients to which we append i-vectors
for speaker adaptation. The network is trained with stochastic gradient descent
for 2~epochs.

In preliminary experiments we trained a separate i-vector extractor for each
different sized subset of training languages. However, results were similar to
training on the pooled set of all 10 high-resource languages, so for expedience
we used the 100-dimensional i-vectors from this pooled training for all reported
experiments. Including i-vectors yielded a small performance gain over not doing
so; we also tried applying \gls{vtln} to the MFCCs for \gls{tdnn} training, but
found no additional benefit.

\subsection{Correspondence autoencoder}
\label{sec:cae}

\begin{figure}
  \includegraphics[width=\columnwidth]{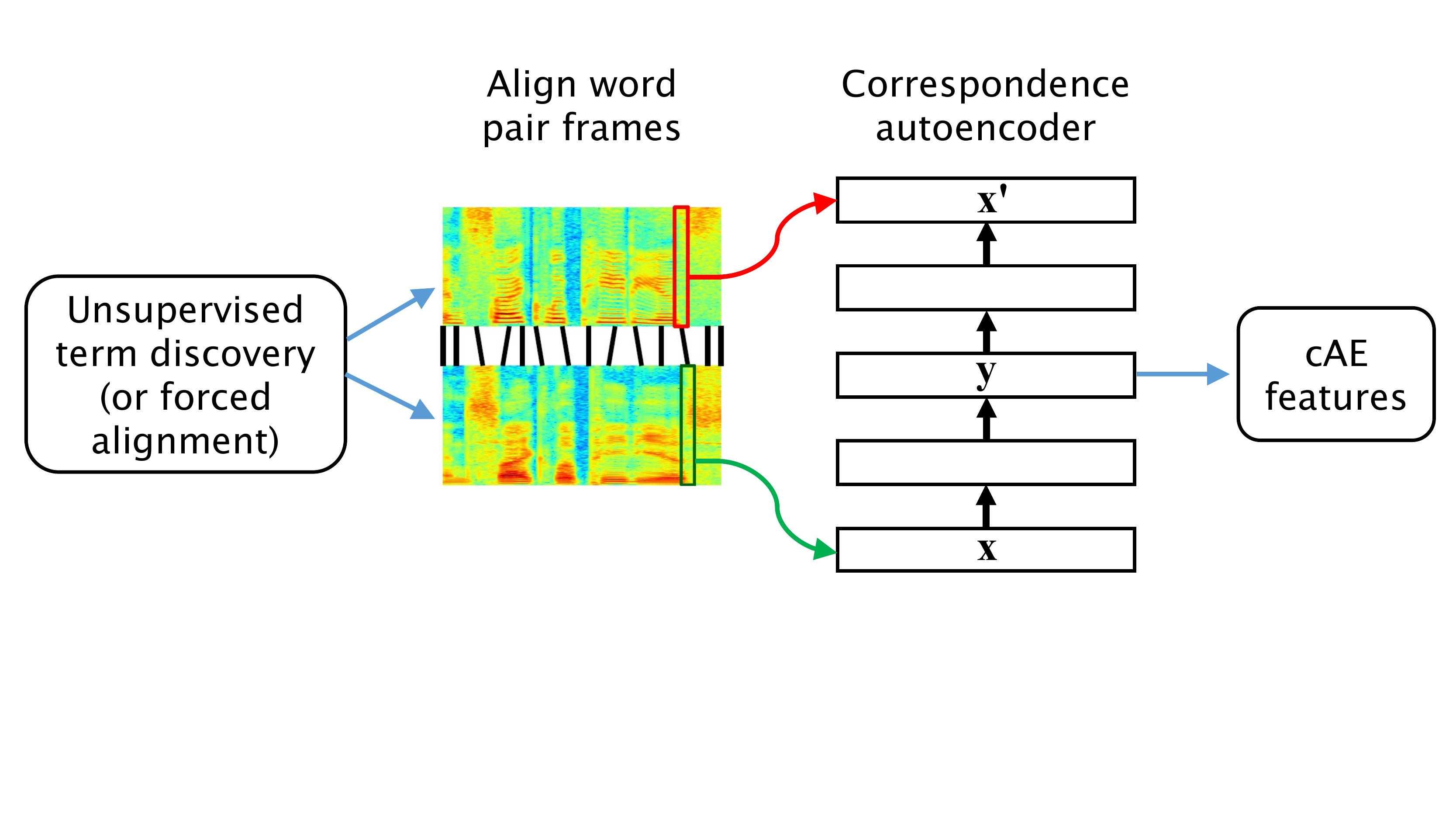}
  \vspace{-5em}
  \caption{Correspondence autoencoder training procedure (see section~\ref{sec:cae}). Parts of this figure due to Herman Kamper, used with permission.}
  \label{f:cae}
\end{figure}

In several experiments we further adapt the baseline features or \gls{bnfs}
using a \gls{cae} network. The \gls{cae} attempts to normalize out
non-linguistic factors such as speaker, channel, gender, etc., using top-down
information from pairs of similar speech segments. Extracting \gls{cae} features
requires three steps, as illustrated in Figure~\ref{f:cae}. First, an
\acrfull{utd} system is applied to the target language to extract pairs of
speech segments that are likely to be instances of the same word or phrase. Each
pair is then aligned at the frame level using \gls{dtw}, and pairs of aligned
frames are presented as the input $\mathbf{x}$ and target output $\mathbf{x'}$
of a \gls{dnn}. After training, a middle layer $\mathbf{y}$ is used as the
learned feature representation.

To obtain the \gls{utd} pairs, we used a freely available \gls{utd}
system\footnote{https://github.com/arenjansen/ZRTools}~\cite{Jansen2011} and
extracted 36k word pairs for each target language. Published results with this
system use PLP features as input, and indeed our preliminary experiments
confirmed that MFCCs did not work as well. We therefore report results using
only PLP or PLP+VTLN features as input to \gls{utd}.

To provide an upper bound on \gls{cae} performance, we also report results using
{\em gold standard} same-word pairs for \gls{cae} training. As in
\cite{Kamper2015,Jansen2013,Yuan2017a}, we force-align the target language data
and extract all the same-word pairs that are at least 5~characters and
0.5~seconds long (between 89k and 102k pairs for each language).

Following~\cite{Renshaw2015,Kamper2015}, we train the \gls{cae}
model\footnote{https://github.com/kamperh/speech\_correspondence} by first
pre-training an autoencoder with eight 100-dimensional layers and a final layer
of size 39 layer-wise on the entire training data for 5~epochs with a learning
rate of $2.5\times10^{-4}$. We then fine-tune the network with same-word pairs
as weak supervision for 60~epochs with a learning rate of $2.5\times10^{-5}$.
Frame pairs are presented to the \gls{cae} using either MFCC, MFCC+VTLN, or BNF
representation, depending on the experiment (preliminary experiments indicated
that PLPs performed worse than MFCCs, so MFCCs are used as the stronger
baseline). Features are extracted from the final hidden layer of the \gls{cae}.

\subsection{Evaluation}
We evaluate all speech features on the same-different task~\cite{Carlin2011}
which tests whether a given speech representation can correctly classify two
speech segments as having the same word type or not. For each word pair in a
pre-defined set $S$ the \gls{dtw} cost between the acoustic feature vectors
under a given representation is computed. Two segments are then considered a
match if the cost is below a threshold. Precision and recall at a given
threshold $\tau$ are defined as

$$ P(\tau) = \frac{M_{\mathrm{SW}}(\tau)}{M_{\text{all}}(\tau)},~~~
R(\tau) = \frac{M_{\mathrm{SWDP}}(\tau)}{|S_{\mathrm{SWDP}}|}$$

where $M$ is the number of \gls{sw}, \gls{swdp} or all discovered matches at
that threshold and $|S_{\mathrm{SWDP}}|$ is the number of actual \gls{swdp}
pairs in $S$. By varying the threshold a precision-recall curve can be computed,
where the final evaluation metric is the \gls{ap} or the area under that curve.
We generate evaluation sets of word pairs for the GlobalPhone development and
test sets as above, from all words that are at least 5~characters and
0.5~seconds long, except that we now also include different-word pairs.

We note that previous work~\cite{Carlin2011,Kamper2015} computed recall with all
\gls{sw} pairs for easier computation because their test sets included a
negligible number of \gls{swsp} pairs. In our case the smaller number of
speakers in the GlobalPhone corpora results in up to 60\% of \gls{sw} pairs
being from the same speaker. We therefore explicitly compute the recall only for
\gls{swdp} pairs to focus the evaluation of features on their speaker
invariance.

As a sanity check, we also provide word error rates (WER) for the \gls{asr}
systems trained on the high-resource languages.

\section{Results}

\subsection{Using target language data only}

Our first set of experiments aims to find the best features that can be
extracted using target language data only. Previous work has shown that
\gls{cae} features are better than MFCCs, especially for cross-speaker word
discrimination \cite{Renshaw2015}, but we know of no direct comparison between
\gls{cae} features and \gls{vtln}, which can also be trained without
transcriptions.

Table~\ref{tab:vtln-results} shows AP results on all target languages for
baseline features, \gls{cae} features learned using raw features as input (as in
previous work), and \gls{cae} features learned using \gls{vtln}-adapted features
as input to either the \gls{utd} system, the \gls{cae}, or both. We find that
\gls{cae} features as trained previously are slightly better than MFCC+VTLN, but
can be improved considerably by applying \gls{vtln} to the input of both
\gls{utd} and \gls{cae} training---indeed, even using gold pairs as \gls{cae}
input applying VTLN is beneficial. This suggests that \gls{cae} training and
VTLN abstract over different aspects of the speech signal, and that both should
be used when only target language data is available.

\begin{table}[t]
    \footnotesize
    \caption{Average precision scores on the same-different task (dev sets),
      showing the effects of applying \gls{vtln} to the input features for the
      \gls{utd} and/or \gls{cae} systems. \gls{cae} input is either MFCC or
      MFCC+VTLN. Topline results (rows 5-6) train cAE on gold standard pairs,
      rather than UTD output. Baseline results (final rows) directly evaluate
      acoustic features without UTD/cAE training. Best unsupervised result in
      bold.}
\label{tab:vtln-results}
  \centering
  \begin{tabular}{@{~~}l l c@{~~~~}c@{~~~~}c@{~~~~}c@{~~~~}c@{~~~~}c@{~~}}
    \toprule
    \multicolumn{1}{c}{\parbox{1cm}{\textbf{UTD\\ input}}} &
    \multicolumn{1}{c}{\parbox{.7cm}{\textbf{cAE \\input}}} & 
    \textbf{ES} &
    \textbf{HA} &
    \textbf{HR} &
    \textbf{SV} &
    \textbf{TR} &
    \textbf{ZH}\\
    \midrule
    PLP &       & 28.6 & 39.9 & 26.9 & 22.2 & 25.2 & 20.4 \\
    PLP & +VTLN & 46.2 & 48.2 & 36.3 & 37.9 & 31.4 & 35.7 \\
    \addlinespace[0.5em]
    PLP+VTLN &       & 40.4 & 45.7 & 35.8 & 25.8 & 25.9 & 26.9 \\
    PLP+VTLN & +VTLN & \textbf{51.5} & \textbf{52.9} & \textbf{39.6} & \textbf{42.9} & \textbf{33.4} & \textbf{44.4} \\
    \midrule
    \textit{Gold pairs} &       & 65.3 & 65.2 & 55.6 & 52.9 & 50.6 & 60.5 \\
    \textit{Gold pairs} & +VTLN & 68.9 & 70.1 & 57.8 & 56.9 & 56.3 & 69.5 \\
    \midrule
    \midrule
    \multicolumn{2}{@{~~}l}{\textit{Baseline:} MFCC}
    & 18.3 & 19.6 & 17.6 & 12.3 & 16.8 & 18.3 \\
    \multicolumn{2}{@{~~}l}{\textit{Baseline:} MFCC+VTLN}
    & 27.4 & 28.4 & 23.2 & 20.4 & 21.3 & 27.7 \\
    \bottomrule
  \end{tabular}
\end{table}

\begin{table}[b]
  \caption{Word error rates of monolingual \gls{sgmm} and 10-lingual TDNN ASR system
    evaluated on the development sets.}
  \label{tab:wer-results}
  \centering
  \begin{tabular}{l c c}
    \toprule
    \multicolumn{1}{c}{\textbf{Language}} & 
    \multicolumn{1}{c}{\textbf{Mono}} &
    \multicolumn{1}{c}{\textbf{Multi}} \\
    \midrule
    BG & 17.5 & 16.9 \\
    CS & 17.1 & 15.7 \\
    DE & 9.6 & 9.3 \\
    FR & 24.5 & 24.0 \\
    KO & 20.3 & 19.3 \\
    \bottomrule
  \end{tabular}
  \hfill
  \begin{tabular}{l c c}
    \toprule
    \multicolumn{1}{c}{} & 
    \multicolumn{1}{c}{\textbf{Mono}} &
    \multicolumn{1}{c}{\textbf{Multi}} \\
    \midrule
    PL & 16.5 & 15.1 \\
    PT & 20.5 & 19.9 \\
    RU & 27.5 & 26.9 \\
    TH & 34.3 & 33.3 \\
    VI & 11.3 & 11.6 \\
    \bottomrule
  \end{tabular}
\end{table}

\subsection{Multilingual training}

\begin{figure*}[t!]
  \includestandalone[width=\textwidth]{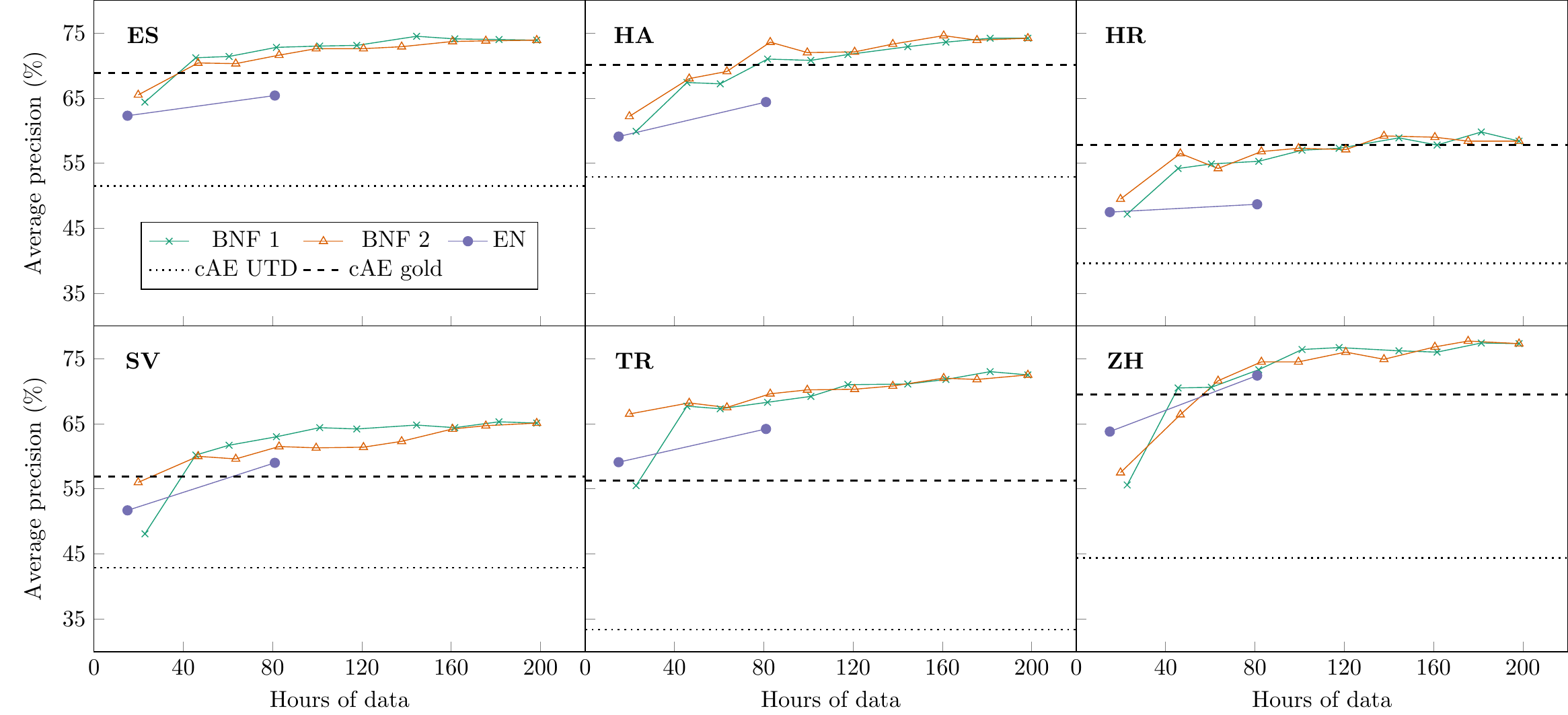}
  \caption{Same-different task evaluation on the development sets for BNFs
    trained on different amounts of data. We compare training on up to
    10~different languages with additional data in one language (English). For
    multilingual training, languages were added in two different orders:
    FR-PT-DE-TH-PL-KO-CS-BG-RU-VI (BNFs 1) and RU-CZ-VI-PL-KO-TH-BG-PT-DE-FR
    (BNFs 2). Each datapoint shows the result of adding an additional language.
    As baselines we include the best unsupervised cAE and the cAE trained on
    gold standard pairs from rows 4 and 6 of Table~\ref{tab:vtln-results}.}
  \label{fig:bnf-results}
\end{figure*}

Table~\ref{tab:wer-results} compares the WER of the monolingual \gls{sgmm}
systems which provide the targets for \gls{tdnn} training to the WER of the
final model trained on all 10~high-resource languages. The multilingual model
shows small but consistent improvements for all languages except Vietnamese.
Ultimately though, we are not so much interested in the performance on typical
\gls{asr} tasks, but in whether \gls{bnfs} from this model also generalize to
zero-resource applications on unseen languages.

Figure~\ref{fig:bnf-results} shows \gls{ap} on the same-different task of
multilingual \gls{bnfs} trained from scratch on an increasing number of
languages in two randomly chosen orders. We provide two baselines for
comparison, drawn from our results in Table~\ref{tab:vtln-results}. Firstly, our
best \gls{cae} features trained with \gls{utd} pairs (from row 4 of
Table~\ref{tab:vtln-results}) are a reference for a fully unsupervised system.
Secondly, the best \gls{cae} features trained with gold standard pairs (from row
6 of Table~\ref{tab:vtln-results}) give an upper bound on the \gls{cae}
performance.

In all 6~languages, even \gls{bnfs} from a monolingual \gls{tdnn} already
considerably outperform the \gls{cae} trained with \gls{utd} pairs. Adding
another language usually leads to an increase in \gls{ap}, with the \gls{bnfs}
trained on 8--10~high-resource languages performing the best, also always
beating the gold \gls{cae}. However, the biggest performance gain is from adding
a second training language---further increases are mostly smaller. The order of
languages has only a small effect, although for example adding other Slavic
languages is generally associated with an increase in \gls{ap} on Croatian,
suggesting that it may be beneficial to train on languages related to the
zero-resource language.

To determine whether these gains come from the diversity of training languages
or just the larger amount of training data, we trained models on the 15~hour
subset and the full 81~hours of the English \gls{wsj} corpus, which corresponds
to the amount of data of four GlobalPhone languages. More data does help to some
degree, as Figure~\ref{fig:bnf-results} shows, but except for Mandarin training
on just two languages (46~hours) already works better.

\begin{table}[t]
  \caption{\Gls{ap} on the same-different task when training
    \gls{cae} on the 10-lingual \gls{bnfs} from above (cAE-BNF) with UTD and
    gold standard word pairs (test set results). Baselines are MFCC+VTLN and the
    cAE models from rows 4 and 6 of Table~\ref{tab:vtln-results} that use
    MFCC+VTLN as input features. Best result without target language supervision
    in bold.}
  \label{tab:cae-results}
  \centering
  \begin{tabular}{l c@{~~~~}c@{~~~~}c@{~~~~}c@{~~~~}c@{~~~~}c@{}}
    \toprule
    \multicolumn{1}{c}{\textbf{Features}} & 
    \textbf{ES} &
    \textbf{HA} &
    \textbf{HR} &
    \textbf{SV} &
    \textbf{TR} &
    \textbf{ZH}\\
    \midrule
    MFCC+VTLN       & 44.1 & 22.3 & 25.0 & 34.3 & 17.9 & 33.4 \\
    cAE UTD         & 72.1 & 41.6 & 41.6 & 53.2 & 29.3 & 52.8 \\
    cAE gold        & 85.1 & 66.3 & 58.9 & 67.1 & 47.9 & 70.8 \\
    \midrule
    10-lingual BNFs & \textbf{85.3} & \textbf{71.0} & \textbf{56.8} & 72.0 & \textbf{65.3} & 77.5 \\
    \midrule
    cAE-BNF UTD     & 85.0 & 67.4 & 40.3 & \textbf{74.3} & 64.6 & \textbf{78.8} \\
    cAE-BNF gold    & 89.2 & 79.0 & 60.8 & 79.9 & 69.5 & 81.6 \\
    \bottomrule
  \end{tabular}
\end{table}

\subsection{cAE results}
Previous work~\cite{Kamper2015} and our baselines in
Table~\ref{tab:vtln-results} show that a fully unsupervised system like a
\gls{cae} generates features that can discriminate between words much better
than standard acoustic features like MFCCs. Is the \gls{cae} also able to
further improve on multilingual \gls{bnfs} which already have a much higher
baseline performance?

We trained the \gls{cae} with the same sets of same-word pairs as before, but
replaced VTLN-adapted MFCCs with the 10-lingual \gls{bnfs} as input features
without any other changes in the training procedure. Table~\ref{tab:cae-results}
shows that the \gls{cae} trained with \gls{utd} pairs is able to slightly
improve on the \gls{bnfs} in some cases, but this is not consistent across all
languages and for Croatian the \gls{cae} features are much worse. The limiting
factor appears to be the quality of the \gls{utd} pairs. With gold standard
pairs, the \gls{cae} features improve in all languages.

\section{Conclusions}
We evaluated multilingual \gls{bnfs} trained on up to 10~high-resource languages
on a word discrimination task in 6~zero-resource languages. These \gls{bnfs}
outperform both standard acoustic features like MFCCs and \gls{cae} features
trained in a fully unsupervised way. We showed that training on multiple
languages helps the \gls{bnfs} and that just training on more data in a single
language does not work as well. While the \gls{cae} is theoretically able to
further improve on the \gls{bnfs}, this does not work in practice if only word
pairs discovered by a \gls{utd} system are available. In future work we would
like to further analyze the complementary nature of \gls{vtln} and \gls{cae}
training and explore the benefits of these multilingual \gls{bnfs} for
down-stream zero-resource applications like speech-to-text translation.

\section{Acknowledgements}
We thank Andrea Carmantini for helping to set up multilingual training for the
GlobalPhone corpus in Kaldi and Herman Kamper for helpful feedback. The research
was funded in part by a James S. McDonnell Foundation Scholar Award.

\bibliographystyle{IEEEtran}
\bibliography{references}

\end{document}